\documentclass[journal]{IEEEtran}
\ifCLASSINFOpdf
\else
\fi

\usepackage{subfigure}
\usepackage{multirow}
\usepackage{multicol}
\usepackage{bigstrut}
\usepackage{booktabs}
\usepackage{comment}
\usepackage{times}
\usepackage{epsfig}
\usepackage{graphicx}
\usepackage{amsmath}
\usepackage{amssymb}
\usepackage{enumitem}
\usepackage{makecell}
\usepackage{bm} 
\usepackage[dvipsnames, svgnames, x11names]{xcolor}
\usepackage[switch]{lineno}
\usepackage{amssymb}
\usepackage{pifont}

\newcommand{\etal}{\textit{et al}.~}

\newcommand{\ieno}{\textit{i}.\textit{e}.}

\newcommand{\egno}{\textit{e}.\textit{g}.} 

\newcommand{\cmark}{\ding{51}}%
\newcommand{\xmark}{\ding{55}}%

\begin{document}
%
\title{Beyond Triplet Loss: Meta Prototypical N-tuple Loss for Person Re-identification}
%
%
%

\author{Zhizheng Zhang, 
        Cuiling Lan,~\IEEEmembership{Member,~IEEE,}
        Wenjun Zeng,~\IEEEmembership{Fellow,~IEEE,} \\
        Zhibo Chen,~\IEEEmembership{Senior Member,~IEEE,}
        and~ Shih-Fu Chang,~\IEEEmembership{Fellow,~IEEE}
        
\thanks{Zhizheng Zhang and Zhibo Chen are with the CAS Key Laboratory of Technology in Geo-Spatial Information Processing and Application System, University of Science and Technology of China, Hefei, Anhui, P.R. China. (E-mail: zhizheng@mail.ustc.edu.cn, chenzhibo@ustc.edu.cn) \\
Cuiling Lan and Wenjun Zeng are with Microsoft Research Asia, Beijing, P.R. China. (E-mail: {culan, wezeng}@microsoft.com) \\
Shih-Fu Chang is with Columbia University, New York, NY. (E-mail: sc250@columbia.edu)
}}

\markboth{Journal of \LaTeX\ Class Files,~Vol.~14, No.~8, August~2015}%
{Shell \MakeLowercase{\textit{et al.}}: Bare Demo of IEEEtran.cls for IEEE Journals}
%



\maketitle

\begin{abstract}
Person Re-identification (ReID) aims at matching a person of interest across images. In convolutional neural network (CNN) based approaches, loss design plays a vital role in pulling closer features of the same identity and pushing far apart features of different identities. In recent years, triplet loss achieves superior performance and is predominant in ReID. However, triplet loss considers only three instances of two classes in per-query optimization (with an anchor sample as query) and it is actually equivalent to a two-class classification. There is a lack of loss design which enables the joint optimization of multiple instances (of multiple classes) within per-query optimization for person ReID. In this paper, we introduce a multi-class classification loss, \ieno, N-tuple loss, to jointly consider multiple ($N$) instances for per-query optimization. This in fact aligns better with the ReID test/inference process, which conducts the ranking/comparisons among multiple instances. Furthermore, for more efficient multi-class classification, we propose a new meta prototypical N-tuple loss. With the multi-class classification incorporated, our model achieves the state-of-the-art performance on the benchmark person ReID datasets.
\end{abstract}


%
\IEEEpeerreviewmaketitle

\section{Introduction}
\label{sec:introduction}

\IEEEPARstart{P}{erson} re-identification (ReID) aims to identify the same persons across images captured at different times, or places, or from different cameras. It has drawn a lot of attention from both academia and industry. The objective of CNN-based person ReID methods is to minimize the feature discrepancies (distances) among the samples with the same identity while maximizing the feature discrepancies (distances) among the samples of different identities to encourage the separation of positive pairs and negative pairs.

Person ReID lies in between image classification \cite{krizhevsky2012imagenet,zheng2017person}, where each identity is treated as one class in the training, and instance retrieval \cite{zheng2017sift}. The identities are available during training while the identities of test images are previously ``unseen" \cite{zheng2016person}. The person ReID test can be considered as a retrieval process. Given a query image, its distances (or similarities) to all the samples in the gallery set will be calculated and ranked to identify the matched images. Given a person sample as the anchor, triplet loss optimizes its distance to one sample of the same identity to be closer than its distance to another sample of a different identity. There are also many works \cite{li2017person,sun2018beyond,yao2019deep,zhai2019defense}, that use the conventional classification loss for feature learning, casting person ReID as a classification problem over all identities in the training set (with each identity taken as a category). Recently, many state-of-the-art works combine both triplet loss and classification loss, which leads to a superior performance to those using only one of them  \cite{wang2018learning,zhang2019densely,zhang2019relation,lawen2019attention,luo2019strong}.

\begin{figure}[t]
	\begin{center}
		\includegraphics[width=.96\linewidth]{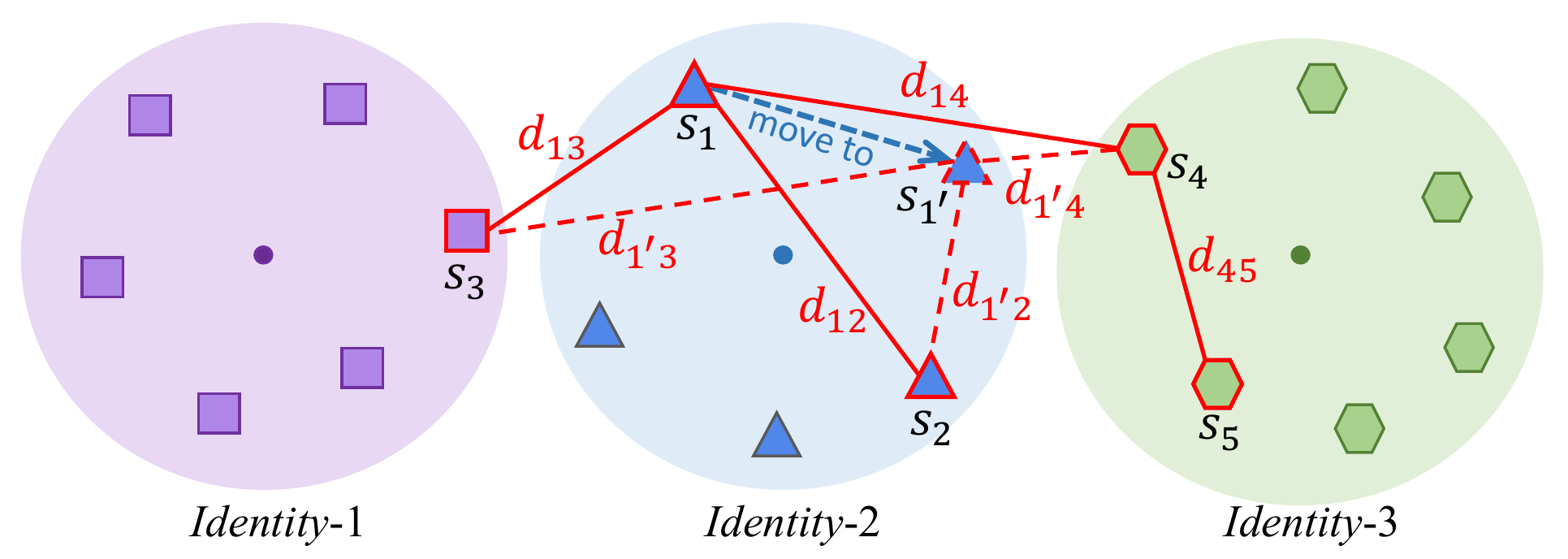}
	\end{center}
	\caption{A toy example of samples of three different identities identified by different shapes (rectangle, triangle or hexagon) in the embedding space. The solid circles denote the identity centers. The samples ($s_1$, $s_2$, $s_3$) constitute a triplet wherein $s_1$ is the anchor, $s_2$ is the positive sample (that has the same ID with the anchor), and $s_3$ is the negative sample (that has a different ID). Similarly, ($s_4$, $s_5$, $s_1$) constitute another triplet.}
	\label{fig:examples}
\end{figure}

Fig. \ref{fig:examples} illustrates the roles triplet loss and classification loss play in optimization and also reveals the limitation of the triplet loss. The conventional classification loss optimizes towards pulling samples closer to their corresponding class centers (\ieno, the coefficients of the fully connected layer of the classifier) and pushing them farther apart from other class centers. This encourages the separation of class centers, which enables a global-wise optimization but does not ensure the reasonableness of the relative order of different instances. As the examples shown in Fig. \ref{fig:examples}, although all the samples can be correctly classified with a conventional classifier (that is implemented by network parameters), the distance between sample $s_1$ and $s_2$ (of the same identity) is still larger than the distance between $s_1$ and $s_3$ (of different identities). This contradicts ReID inference which performs ranking among instances.

The triplet loss directly optimizes the relative order of three samples/instances from two identities. As illustrated in Fig.~\ref{fig:examples}, a triplet consists of an anchor sample (\egno, $s_1$), a positive sample (\egno, $s_2$) that has the same identity as the anchor, and a negative sample with different identity (\egno, $s_3$). The triplet loss aims to reduce the distance ($d_{12}$) of the positive pair and enlarge the distance of the negative pair ($d_{13}$) to make $d_{13}$ larger than $d_{12}$ in the embedding space. However, when we look into two triplets ($s_1$, $s_2$, $s_3$) and ($s_4$, $s_5$, $s_1$) respectively, their optimization directions may contradict each other and thus render inferior results. For example, sample $s_1$ may move to a new position $s_1^{'}$ as a result of optimizing for the first triplet. However, due to lack of efficient interaction with the second triplet, the optimized new position $s_1^{'}$ would make the second triplet worse. For the anchor $s_4$, the distance with the negative sample $s_1$ becomes smaller. Thus, considering multiple instances (\egno, 5 samples) through several (\egno, 2) independent triplets is hard to provide powerful constraints.

To address the above limitation of triplet loss for person ReID, we rethink the loss designs and propose to exploit $N$-tuple loss to jointly optimize multiple instances (from multiple classes) within a per-query optimization. This enables the joint comparison across multi-class instances which better matches the nature of ReID inference, \ieno, distance-based ranking across many images in the gallery set. We validate the effectiveness of the N-tuple loss on several benchmark datasets. Moreover, we propose a lifted variant of $N$-tuple loss, meta prototypical N-tuple (MPN-tuple) loss for more efficient multi-class classification. In the MPN-tuple loss, we employ a meta-predictor to learn the category-specific prototype from instances as the classifier for multi-class classification, which is optimized in an end-to-end manner.

We summarize our main contributions as follows: 

\begin{itemize}[leftmargin=*,noitemsep,nolistsep]

\item We introduce the N-tuple loss, under a unified loss formulation, to person ReID, which jointly optimizes over multiple instances from multiple classes for each query sample. This in fact aligns better with the ReID test/inference process but is under-explored in previous ReID works.

\item We propose a variant of N-tuple loss, \ieno, meta prototypical N-tuple (MPN-tuple) loss, towards more efficient multi-class classification optimization for person ReID.

\item As a minor contribution, we conduct a systematic empirical study by revisiting the design choices of both triplet loss and conventional classification loss as well as their combinations. We use the best practices to serve as our baseline.

\end{itemize}

While simple, our scheme achieves the state-of-the-art performance, outperforming those of previous loss designs by a large margin.
We hope our scheme could serve as a new strong baseline which benefits the ReID community in the future.

\section{Related Works}

\noindent\textbf{Person Re-identification.}
Many efforts have been made for representative feature learning from the perspectives of network or loss designs for person ReID.

Some approaches exploit multi-granularity feature representation to capture both global and local features for person ReID \cite{bai2017deep,wang2018learning,li2017person,wei2017glad,su2017pose,zhao2017spindle}. Some others introduce attention mechanisms for discriminative feature learning, \cite{li2018harmonious,wang2018mancs,zhang2019relation,chen2019mixed,song2018mask}. To tackle the challenges of diverse viewpoints and poses, many works exploit some auxiliary semantics (\egno segmentation \cite{song2018mask}, human parsing \cite{kalayeh2018human}, pose \cite{su2017pose,zhao2017spindle}, dense semantics \cite{zhang2019densely,jin2019semantics}) to address the misalignment problem in person ReID. 

For loss designs in person ReID,  the classification loss is widely used, with the total number of classes being the number of identities in the training set \cite{li2017person,sun2018beyond,yao2019deep,zhai2019defense}. In the early years, some works employ the contrastive loss \cite{varior2016gated,varior2016siamese,wang2016joint} and verification loss \cite{li2014deepreid,ahmed2015improved} to optimize the instances. By introducing relative distance order between the positive sample pair and the negative sample pair, triplet loss and its variants prevail \cite{khamis2014joint,paisitkriangkrai2015learning,cheng2016person,hermans2017defense,zhou2017point,yu2018hard,almazan2018re,yuan2019defense} for person ReID. Hermans \etal introduce a batch-level hard triplet mining, which selects the hardest positive and the hardest negative samples within a batch for optimization~\cite{hermans2017defense}. Chen \etal propose the quadruplet loss which is built based on triplet loss and additionally pushes away negative pairs from positive pairs w.r.t. different probe images~\cite{chen2017beyond}. \emph{These losses do not jointly consider the relative order of multiple instances (\egno $>$3) in per-query optimization.} 

Most recent works combine conventional classification loss and triplet-based loss in optimization for a higher performance \cite{wang2018learning,zhang2019densely,zhang2019relation,fang2019bilinear,lawen2019attention,luo2019strong}. In this work, we introduce N-tuple loss and further develop its improved version for person ReID, which jointly optimizes multiple instances of multiple identities for a given query. This aligns better with the person ReID test/inference and thus leads to a superior performance.

\noindent\textbf{Metric Learning.}
The study on metric learning \cite{roweis2004neighbourhood,weinberger2006distance} stemmed from the era before deep learning. It has been an indispensable part of deep learning in the form of loss design for many applications, such as person ReID \cite{varior2016siamese,hermans2017defense,chen2017beyond}, face recognition \cite{schroff2015facenet}, few-shot learning \cite{vinyals2016matching,snell2017prototypical}. As one of the most commonly used pairwise losses, contrastive loss is investigated in \cite{hadsell2006dimensionality}. CenterLoss \cite{wen2016discriminative} calculates class-specific center in the embedding space and explicitly optimizes the intra-class compactness by explicitly minimizing the Euclidean distances between samples and their corresponding class centers. However, it leaves the inter-class distances under-considered. Triplet-based losses setup an anchor and pull the distance of the positive pair to be smaller than the negative pair. To guarantee the effectiveness of selected triplets, batch hard mining \cite{hermans2017defense} and soft margin \cite{hermans2017defense,lawen2019attention} are widely used. Triplet loss is also improved in its speed and robustness by generalizing from optimizing instance-to-instance distances to optimizing instance-to-centroid distances towards its upper bound \cite{do2019theoretically}. N-tuple loss pushes $N-2$ negative samples and pulls the positive pair all at once \cite{sohn2016improved}. Being conceptually interesting, however, such joint ordering of multiple classes is under-explored in ReID. In this paper, we raise this under-explored issue in current loss designs for ReID and highlight that multi-class classification has an important impact on its performance, hopefully enabling other researchers in the field to leverage the full potential of multi-class joint optimization.


\section{MPN-tuple Loss for Person ReID}
To better understand the prevalent loss designs in person ReID, we first revisit the roles different loss designs (\ieno, classification loss and triplet loss) play in optimizing person ReID models. To remedy the limitation of adopting triplet loss and enable more effective feature learning, we introduce N-tuple loss to this field which aligns better with the test process of person ReID. This facilitates the joint comparisons with multiple instances per query sample rather than only with two samples (in triplet loss). Moreover, we propose a new variant of N-tuple loss, named meta prototypical N-tuple (MPN-tuple) loss, to further promote the metric learning.

\subsection{Revisiting Loss Designs Under a Unified View}

We revisit different loss designs under a unified formulation. We show that the triplet loss can be treated as optimizing a two-class classification where the query/anchor sample is compared with two reference samples (a positive sample and a negative sample), and the probability of being classified as the positive class is maximized in the optimization.

\noindent\textbf{A Unified Loss Formulation.}
To facilitate the understanding, we formulate the loss designs (for classifying the anchor as the $j^{th}$ class) from a unified classification view as:

\begin{equation}
\begin{aligned}
\!\mathcal{L}_{unified}^{i}\! &=\!- {\rm{log}}\frac{{\rm{exp}}(\frac{1}{\tau}\mathcal{S}(\boldsymbol{x}_a,\boldsymbol{c}_{i}))}{\sum_{k=1}^{C} {\rm{exp}}(\frac{1}{\tau}\mathcal{S}(\boldsymbol{x}_a,\boldsymbol{c}_k))} \! \\
&= \!{\rm{log}} \left(1\! +\!\frac{ \sum_{k\neq j}^{C} {\rm{exp}}(\frac{1}{\tau}\mathcal{S}(\boldsymbol{x}_a,\boldsymbol{c}_k))}{{\rm{exp}}(\frac{1}{\tau}\mathcal{S}(\boldsymbol{x}_a,\boldsymbol{c}_{j}))} \right),
\label{eq:unifiedloss}
\end{aligned}
\end{equation}
where $\mathcal{S}(\cdot,\cdot)$ denotes the similarity between two feature vectors/nodes, $\boldsymbol{x}_a \in \mathbb{R}^d$ denotes the feature vector (of $d$ dimensions) of an anchor~(query) sample $a$ to be classified/matched, and $\boldsymbol{c}_i \in \mathbb{R}^d$ denotes the weight vector corresponding to the class of $\boldsymbol{x}_a$ in the classifier. $\tau$ denotes a temperature parameter. We can write the weight matrix as $\mathbf{W}_b = [\boldsymbol{c}_1, \boldsymbol{c}_2, \cdots, \boldsymbol{c}_C] \in \mathbb{R}^{d\times C}$. Generally, $\boldsymbol{c}_k$ with $k=1,2,\cdots,C$ can be considered as the \emph{reference nodes} for classification/matching where each node acts as the feature representation of the class center of a category. $C$ denotes the number of the reference classes or instances. The more similar between $\boldsymbol{x}_a$ and a reference $\boldsymbol{c}_j$, the higher of the probability that they belong to the same class. \emph{Minimizing this loss is to maximize the probability of correct classification.}

The Softmax function in Eq.(\ref{eq:unifiedloss}) plays the role of normalization and enabling the interaction between the query sample and the reference nodes.
For the query sample $\boldsymbol{x}_a$, it optimizes the similarities/distances between the query and the reference nodes, enabling the joint comparisons among these pairs.

This unified formulation can be instantiated to different loss designs. Here, we discuss the widely used conventional classification loss and triplet loss in person ReID in detail.

\noindent\textbf{Conventional Classification Loss.}
In the unified formulation, when $C$ is the total number
of classes/identities in the training set, and the weight matrix $\mathbf{W}_b = [\boldsymbol{c}_1, \boldsymbol{c}_2, \cdots, \boldsymbol{c}_C] \in \mathbb{R}^{d\times C}$ is composed of the learned weights of the Fully-Connected (FC) layer of a classifier and $\mathcal{S}(\cdot,\cdot)$ is calculated by the inner product, the loss for the sample $\boldsymbol{x}_a$ becomes the conventional classification loss as:
\begin{equation}
\!\mathcal{L}_{cls}\!=\!- {\rm{log}}\frac{{\rm{exp}}(\frac{1}{\tau}\boldsymbol{c}_{i}^{\rm{T}}\boldsymbol{x}_a)}{\sum_{k=1}^{C} {\rm{exp}}(\frac{1}{\tau}\boldsymbol{c}_{k}^{\rm{T}}\boldsymbol{x}_a))},
\label{eq:clsloss}
\end{equation}
which denotes the negative logarithm of the probability of classifying the sample $\boldsymbol{x}_a$ into the $i^{th}$ class. $C$ is the total number of classes/identities in the training set. Each reference node $\boldsymbol{c}_k$ is a learned weight vector (\ieno, network parameters), that plays the role of ``class center". The classification probability for the sample $\boldsymbol{x}_a$ is obtained by comparing the similarities/distances between $\boldsymbol{x}_a$ and all the ``class centers".

The conventional classification loss as formulated in Eq.(\ref{eq:clsloss}) is widely used in person ReID. It optimizes the similarities between the query instance (to be classified) and all ``class centers", which plays a role of globally optimizing the ``class centers". The classification loss thus inclines to improve class discrimination but lacks the adaptability in the instance level. As illustrated by the toy examples in Fig. \ref{fig:examples}, even when all samples are already correctly classified based on the similarity to the class centers, it is still questionable to determine the relative order of different instances for a given query based on their distances. For example, for the query sample $s_1$, its distance to the positive sample $s_2$ is even larger than that to the negative sample $s_3$. \emph{Conventional classification loss lacks the capability of optimizing the relative order of instance pairs.}

\noindent\textbf{Triplet Loss.} 
Triplet loss explicitly optimizes the distances/similarities of a positive instance pair and a negative instance pair for a given query. It remedies the above discussed limitation of the conventional classification loss. The vanilla version of triplet loss encourages the similarity between the anchor/query and a positive sample to be larger than the similarity between this anchor and a negative sample by a hard margin $m$ as below:

\begin{equation}\label{eq:triplet-H}
\mathcal{L}_{triplet}=[m+\mathcal{S}(\boldsymbol{x}_a, \boldsymbol{x}^-) - \mathcal{S}(\boldsymbol{x}_a, \boldsymbol{x}^+)]_+,
\end{equation}
where $\boldsymbol{x}_a$, $\boldsymbol{x}^+$, and $\boldsymbol{x}^-$ denote an anchor sample, a positive sample (that has the same identity as $\boldsymbol{x}_a$), and a negative sample (that has a different identity to $\boldsymbol{x}_a$), respectively. $[\cdot]_+\!=\!max(\cdot, 0)$. The soft margin variant of triplet loss has been demonstrated to be more effective for person ReID \cite{hermans2017defense,lawen2019attention} than its other variants. It replaces the hinge function $[m+ \cdot\,]_+$ by softplus function ${\rm{log}}(1+{\rm{exp}}(\cdot))$ which decays exponentially instead of having a hard cut off. It is defined as:
\begin{equation}\label{eq:triplet-S}
\mathcal{L}_{triplet}={\rm{log}}(1+{\rm{exp}}(\mathcal{S}(\boldsymbol{x}_a, \boldsymbol{x}^-)-\mathcal{S}(\boldsymbol{x}_a, \boldsymbol{x}^+)),
\end{equation}
which is equivalent to:
\begin{equation}\label{eq:triplet-C}
\begin{aligned}
\mathcal{L}_{triplet}&=\mathrm{log}(1+\mathrm{exp}(\mathcal{S}(\boldsymbol{x}_a, \boldsymbol{x}^-)/\mathrm{exp}(\mathcal{S}(\boldsymbol{x}_a, \boldsymbol{x}^+))\\&=-\mathrm{log}\frac{\mathrm{exp}(\mathcal{S}(\boldsymbol{x}_a, \boldsymbol{x}^+))}{\mathrm{exp}(\mathcal{S}(\boldsymbol{x}_a, \boldsymbol{x}^+))+\mathrm{exp}(\mathcal{S}(\boldsymbol{x}_a, \boldsymbol{x}^-))}.
\end{aligned}
\end{equation}
Comparing Eq.(\ref{eq:unifiedloss}) and Eq.(\ref{eq:triplet-C}), triplet loss with soft margin is actually an instantiation of the unified classification loss (see Eq.(\ref{eq:unifiedloss})), by setting $C$=$2$ and taking the positive sample $\boldsymbol{x}^+$ and the negative sample $\boldsymbol{x}^-$ as reference nodes. The loss maximizes the probability of the sample $\boldsymbol{x}_a$ to be classified into the class corresponding to the positive sample $\boldsymbol{x}^+$, by encouraging a higher similarity with the positive sample and a lower similarity with the negative sample. We thus view the soft-margin triplet loss as a classification loss over two categories/identities (\ieno, a two-class classification loss). It optimizes instance-to-instance distances with two instances as the reference nodes. A temperature factor $1/\tau$ is multiplied over the similarity function.

For triplet loss, only three instances of two classes are jointly optimized at once in a per-query optimization. In a min-batch, multiple triplets are usually sampled to calculate the batch-level triplet losses. One may argue that when we look at multiple triplets (\egno, two), multiple instances are also ``jointly'' optimized. Actually, this viewpoint is questionable. For example, in the case of two triplets, there is a lack of valid interaction between them. \emph{The optimization directions of the two triplets may contradict each other as illustrated in Fig. \ref{fig:examples} (analyzed in Section \ref{sec:introduction}), leading to inferior optimization.}

\subsection{N-tuple Loss} 

We propose to jointly optimize multiple instances in a per query optimization to address the limitation of triplet loss. Particularly, we increase the number of instances jointly considered in per-query optimization, enabling the comparisons of a query sample with more samples. This better aligns with ReID inference, where a query sample is to be compared with the samples in the gallery set. Although simple and intuitive in concept, this is overlooked in the person ReID literature.

We introduce N-tuple loss to enable the joint optimization of multiple instances (from more than two identities). N-tuple loss \cite{sohn2016improved} allows the interaction of an anchor/query sample with multiple samples from multiple (more than two) different classes. Given an anchor sample $\boldsymbol{x}_a$, N-tuple loss is defined as:
\begin{equation}
\begin{aligned}
\mathcal{L}_{Ntuple}\!&=\!- {\rm{log}}\frac{{\rm{exp}}(\frac{1}{\tau}\mathcal{S}_{a}^{+})}{{\rm{exp}}(\frac{1}{\tau}\mathcal{S}_{a}^{+}\!+\!\sum_{k=1}^{N-2} {\!\!\rm{exp}}(\frac{1}{\tau}\mathcal{S}_{a,k}^{-})}, \\
\mathcal{S}_{a}^{+}&=\mathcal{S}(\boldsymbol{x}_a,\boldsymbol{x}^+), \quad\quad \mathcal{S}_{a,k}^{-}=\mathcal{S}(\boldsymbol{x}_a,\boldsymbol{x}_k^-),
\label{eq:N-tuple}
\end{aligned}
\end{equation}
where the $N$ of a N-tuple refers to the number of elements in this tuple including one anchor sample $\boldsymbol{x}_a$, one positive sample and $N$-2 negative nodes. $\boldsymbol{x}_k^-$, $k=1,\cdots,N-2$ corresponds to $N-2$ negative samples of $N-2$ different classes, and $\boldsymbol{x}^+$ corresponds to a positive sample (same identity as the anchor/query sample). Compared to the soft-margin triplet loss in Eq.(\ref{eq:triplet-C}), the difference in Eq.(\ref{eq:N-tuple}) is that multiple negative samples instead of one are used. When $N=3$, it degenerates to the soft-margin triplet loss. N-tuple loss actually corresponds to a multi-class classification and optimizes instance-to-instance distances with $N-1$ instances as reference nodes and one instance as query. \emph{The joint optimization of multi-instances promotes the relative order of distances among many instance pairs which matches ReID inference well.}

\subsection{Proposed Meta Prototypical N-tuple Loss}

Based on the multi-class classification loss, \ieno, N-tuple loss, we further propose a \emph{Meta Prototypical N-tuple} loss (abbreviated as MPN-tuple loss) for effective optimization in person ReID. In N-tuple loss, feature vectors of instances themselves are taken as the reference nodes (comprising a classifier) to perform multi-class classification for multiple instances. Here, we propose to learn better reference nodes from instance features via a trainable meta-learner. First, we incorporate a trainable meta-learner as a predictor to predict the category-specific reference node from each instance in a mini-batch. Second, when the number of classes jointly taken into account increases in per-query optimization, the total number of tuples increases exponentially and becomes intractable quickly. We average the reference nodes of the same identity within a mini-batch to be the final reference node in the N-tuple optimization. The advantages lie in two aspects. 1) The feature representation of one instance might be affected by the noise of this instance. In contrast, the averaged result across multiple instances is more robust to be a reference node. 2) This reduces the number of N-tuples to make it trackable especially when $N$ is large.

Particularly, instead of directly using the features of sampled instances as the reference nodes (as in N-tuple loss), we propose to employ a meta learner by using a mapping subnet $\phi(\cdot)$ for obtaining the reference nodes based on the refined instance features. We define $\phi(\boldsymbol{x}_k)$ as
\begin{equation}
\phi(\boldsymbol{x}_k)=W_2({\rm{BN}}(W_1(\boldsymbol{x}_k))),
\label{eq:subnet}
\end{equation}
where $\phi(\cdot)$ is implemented by two Fully-Connected (FC) layers with a Batch Normalization (BN) layer, $W_1\!\in\!\mathbb{R}^{\frac{d}{s} \times d}$, $W_2\!\!\in\!\!\mathbb{R}^{d\times \frac{d}{s}}$, wherein $s$ is an integer which controls the dimension reduction ratio and we experimentally set it to 8. The dimension reduction here is inspired by the squeeze and excitation operations in \cite{hu2018squeeze}, which reduces the number of parameters to make the optimization easier. Here, we define the similarity function $\mathcal{S}$ as the cosine similarity of the two input vectors as
\begin{equation}
\mathcal{S}(\boldsymbol{x}_a, \phi(\boldsymbol{x}_k)) = (\phi(\boldsymbol{x}_k)^{\rm{T}} \boldsymbol{x}_a)/(\|\phi(\boldsymbol{x}_k)^{\rm{T}}\|\cdot\|\boldsymbol{x}_a\|).
\label{eq:similarity}
\end{equation}
For an anchor $\boldsymbol{x}_a$, we define the Meta N-tuple loss as:
\begin{equation}
\begin{aligned}
\mathcal{L}_{MNtuple}\!&=\!-\!{\rm{log}}\frac{{\rm{exp}}(\frac{1}{\tau}\mathcal{S}_{a}^{\phi(+)})}{{\rm{exp}}(\frac{1}{\tau}\mathcal{S}_{a}^{\phi(+)})\!+\!\sum_{k=1}^{N-2}\!{\rm{exp}}(\frac{1}{\tau}\mathcal{S}_{a,k}^{\phi(-)})}, \\
\mathcal{S}_{a}^{\phi(+)}&=\mathcal{S}(\boldsymbol{x}_a,\phi(\boldsymbol{x}^+)), \ \quad \mathcal{S}_{a,k}^{\phi(-)}=\mathcal{S}(\boldsymbol{x}_a,\phi(\boldsymbol{x}_k^-)).
\label{eq:Meta-N-tuple}
\end{aligned}
\end{equation}
Essentially, the introduction of $\phi(\cdot)$ enables more effective metric learning, where $\phi(\cdot)$ learns to map extracted features to the refined representations in a trainable way. The optimized feature representations as the reference nodes can then further guide the optimization of features (before the mapping).

For $K$ samples ($\boldsymbol{x}_{c,j}$, with $j=1,\cdots, K$) of the class $c$, we build its prototype reference node by averaging their mapped features as $\widehat{\phi_c} = \frac{1}{K} \sum_{j=1}^K  \phi(\boldsymbol{x}_{c,j})$. 
For an anchor sample $\boldsymbol{x}_a$, we define the Meta Prototypical N-tuple loss as:
\begin{equation}
\begin{aligned}
\mathcal{L}_{MPNtuple}&=\!-\!{\rm{log}}\frac{{\rm{exp}}(\frac{1}{\tau}\mathcal{S}_{a}^{\widehat{\phi^+}})}{{\rm{exp}}(\frac{1}{\tau}\mathcal{S}_{a}^{\widehat{\phi^+}}) \!+\! \sum_{k=1}^{N-2}\!{\rm{exp}}(\frac{1}{\tau}\mathcal{S}_{a,k}^{\widehat{\phi^-}})}, \\
\mathcal{S}_{a}^{\widehat{\phi^+}}&=\mathcal{S}(\boldsymbol{x}_a,\widehat{\phi^+}), \qquad\quad \mathcal{S}_{a,k}^{\widehat{\phi^-}}=\mathcal{S}(\boldsymbol{x}_a,\widehat{\phi_k^-}).
\label{eq:Meta-N-tuple}
\end{aligned}
\end{equation}
where ${\widehat{\phi^+}}$ denotes the prototype reference node obtained from the positive samples (corresponding to the same class as the anchor sample) while ${\widehat{\phi_k^-}}$ denotes that for the $k^{th}$ negative class. We will demonstrate the effectiveness of the proposed MPN-tuple loss for person ReID in the experiment section.

As described above, the prototype reference nodes in our proposed method are obtained through averaging the refined feature representations inferred by the meta-learner. Here, the meta-learner can be formulated as a trainable function $\phi(\cdot)$ as mentioned before, which aims to predict the class centers adaptively. Thus, the class-specific prototypes in Eq.(\ref{eq:Meta-N-tuple}) are updated dynamically as the feature learning so that they are robust to data noises and also compact as shown in the histogram plotting of the following experiment part. We note that the class centers are also adopted in CenterLoss \cite{wen2016discriminative} to regularize each sample to approach the corresponding updated center of the same class for minimizing the intra-class distances. However, it leaves the distances of negative pairs under-considered. In contrast, our proposed MPN-tuple loss enables a joint multi-class instance optimization to optimize the similarities of both positive pairs and negative pairs simultaneously for better metric learning. The testing scenario of person ReID is to match a query sample with the samples of the same identity in the gallery set and avoid it to be matched to different identities. Thus, the MPN-tuple loss aligns better with the optimization objective of this task.

\section{Experiments}

\subsection{Datasets and Evaluation Metrics}

\noindent\textbf{Datasets}. We evaluate our methods using three widely-used person ReID datasets: \ieno, CUHK03 \cite{li2014deepreid}, Market1501 \cite{zheng2015scalable}, DukeMTMC-reID \cite{zheng2017unlabeled} and the large-scale MSMT17 \cite{wei2018person}.

\textbf{CUHK03} \cite{li2014deepreid} consists of 1,467 pedestrians. This dataset provides both manually labeled bounding boxes from 14,096 images and DPM-detected bounding boxes from 14,097 images. We adopt the new training/testing protocol following \cite{zhong2017re, zheng2018pedestrian, he2018recognizing}. In this protocol, 767 identities are used for training and the remaining for testing. We only show the evaluation results for the labeled setting (L) while the detected setting (D) presents a similar trend.

\textbf{Market-1501}~\cite{zheng2015scalable} contains 12,936 images of 751 identities for training and 19,281 images of 750 identities for testing, which are captured by 6 cameras.

\textbf{DukeMTMC-reID}~\cite{zheng2017unlabeled} has 36,411 images, where 702 identities are used for training and 702 identities for testing. They are captured by 8 cameras. 

\textbf{MSMT17}~\cite{wei2018person} contains 126,441 images, where 1,041 identities and 3,060 identities are used for training and testing respectively. They are captured by 15 cameras. 

\noindent\textbf{Evaluation Metrics}. We follow the common practices to use Rank-1 (R1) and mean average precision (mAP) for evaluating person ReID models.

\subsection{Implementation Details}
\label{subsec:implementations}

We perform the empirical study of combining triplet loss and conventional classification loss and evaluate our proposed MPN-tuple loss on the regular supervised person ReID. Besides, we further verify the robustness of MPN-tuple loss on visible-infrared and cloth-changing person ReID. Here, we introduce the configurations of the main body of our experiments in this section, leaving the detailed introduction of experiment configurations for visible-infrared and cloth-changing person ReID in the corresponding sub-sections.

\noindent\textbf{Network Settings}. We follow the common practices in ReID \cite{almazan2018re,zhang2019densely,luo2019strong} and take ResNet-50 \cite{he2016deep} to build our baseline network for effectiveness validation. Similar to \cite{sun2017beyond,zhang2019densely}, we remove the last spatial down-sampling operation in the conv5\_x block of ResNet-50. 
Unless otherwise stated, similar to \cite{pan2018two}, we add Instance Normalization to the first three blocks (conv2\_x-conv4\_x) to enhance model's generalization ability \cite{jia2019frustratingly}, which is found effective in improving the performance because the identities during testing are unseen (different from the training identities). In detail, following IBN-a in \cite{pan2018two}, we replace half of the channels of Batch Normalization (BN) by Instance Normalization (IN) for the first BN layer of the residual blocks within $conv2\_x$, $conv3\_x$ and $conv4\_x$, and leave other BN layers in other positions intact. 

On top of the spatially pooled feature (2048 dimensions) of ResNet-50, a Batch Normalization (BN) layer is added to obtain the ReID feature vector $\boldsymbol{x} \in \mathbb{R}^{1024}$ and a followed Fully Connected (FC) layer is employed as the classifier for adding the conventional classification loss. In our studies, the multi-class classification losses are added on the ReID feature vector $\boldsymbol{x}$ by default. For this loss, similar to \cite{wang2017normface}, the temperature parameter $\tau$ is a learnable  parameter. In our implementation, for the purpose of numerical stability, we learn a weight parameter $s$ where $s= 1/\tau$ instead of $\tau$. Note that we do not employ re-ranking \cite{zhong2017re} in all our experiments.

\noindent\textbf{Training}. We use the commonly used data augmentation strategies of random cropping \cite{wang2018resource}, horizontal flipping, and random erasing \cite{zhong2020random}. The input resolution is set to 384$\times$128. Each batch includes $B\,$=$\,P$$\times$$K$ images. $P$ and $K$ denote the number of different persons (identities) and the number of different images per person, respectively. We perform the experiments with $P\,$=$\,16$, $K\,$=$\,4$ using one GPU card. In a batch, the total number of triplets is $T\,$=$\,C_P^N \cdot K^N N(K-1)\,$=$\,11520$. Even for batch hard mining triplet, the similarities for all the sample pairs in a batch need to be calculated for the selection of the hard triples so that it actually has similar training complexity as using all the triplets. But, as the number of classes increases in N-tuple loss, the total number of tuples increases exponentially which quickly becomes intractable. Therefore sampling the tuples is desirable to limit the complexity.

We initialize the ResNet-50 \cite{he2016deep} backbone network with ImageNet~\cite{deng2009imagenet} pre-trained weights and train the ReID network for 600 epochs in total. An epoch means that all identities in the entire training dataset are traversed. We adopt Adam optimizer with the momentum of 0.9 and the weight decay of $5 \times 10^{-4}$. During the training, we first warm up with a linear growth learning rate from $8 \times 10^{-6}$ to $8 \times 10^{-4}$ for 20 epochs and the learning rate is decayed by a factor of 0.5 for every 60 epochs. Unless otherwise specified, we train the entire network in an end-to-end manner.

\noindent\textbf{Testing}. In the testing, for a given person image, we take the feature vector that is obtained by adopting global average pooling in spatial on the feature map extracted by the ResNet-50 backbone as the final representation of this sample. We calculate the cosine similarities over different person images and perform ranking/retrieval based on such calculated similarities.

\begin{table*}[t]
  \centering
  \caption{A case study for combining triplet loss (Tri.) and conventional classification loss (Cls.). \emph{Distance} denotes the similarity metric used for triplet loss. \emph{HardMining} denotes whether batch hard mining is used in triplet loss.}
    \begin{tabular}{c|c|c|c|c|c|c|c|c|c|c}
    \Xhline{0.8pt}
    \multirow{2}[0]{*}{Loss} & \multirow{2}[0]{*}{Distance} & \multirow{2}[0]{*}{HardMining} & \multicolumn{2}{c|}{CUHK03(L)} & \multicolumn{2}{c|}{Market1501} & \multicolumn{2}{c|}{DukeMTMC} & \multicolumn{2}{c}{MSMT17} \\
    \cline{4-11}
          &       &    & R1   & \,\,mAP\,\,   & R1   & \,\,mAP\,\,   & R1   & \,\,mAP\,\,   & R1   & \,\,mAP\,\, \\
    \hline
    Cls.  &   -    &   -    & 67.2  & 63.6  & 94.1  & 83.5  & 85.6  & 73.8  & 72.7  & 46.8 \\
    \hline
    Tri.  & Euclidean & yes   & 79.4  & 75.0  & 94.0  & 84.8  & 86.9  & 74.6  & 73.8  & 49.8 \\
    Tri.  & Euclidean & no    & 63.7  & 60.2  & 87.6  & 74.0  & 75.9  & 59.0  & 55.0  & 33.0 \\
    Tri.  & Cosine & yes   & 63.3  & 57.5  & 83.1  & 65.3  & 81.6  & 66.4  &  33.2 &  25.6 \\
    Tri.  & Cosine & no    & 43.1  & 40.6  & 68.8  & 50.2  & 65.0  & 46.2  &  25.2  & 12.5 \\
    \hline
    Tri.+Cls. & Euclidean & yes   & 79.6  & 75.8  & 94.9  & 86.6  & 87.3  & 76.7  & 78.8  & 56.0 \\
    Tri.+Cls. & Euclidean & no    & 74.3  & 70.6  & \textbf{95.0}  & 86.9  & 87.3  & 77.1  & 77.6  & 53.9 \\
    Tri.+Cls. & Cosine & yes   & 80.7  & 76.4  & 94.6  & 86.9  & 88.8  & 77.9  & 78.6  & 54.5 \\
    Tri.+Cls. & Cosine & no    & \textbf{81.8}  & \textbf{78.2}  & 94.7  & \textbf{87.3}  & \textbf{88.7}  & \textbf{78.3}  & \textbf{79.8}  & \textbf{56.2} \\
    \Xhline{0.8pt}
    \end{tabular}%
  \label{tab:tri_cls}%
\end{table*}%

\subsection{Empirical Study of Triplet Loss and Conventional Classification Loss for Person ReID}

The joint use of the triplet loss and the conventional classification loss achieves superior performance and is predominant in ReID. For triplet loss, there is a lack of comprehensive study on its design choices and corresponding effectiveness. In this section, we study the good practice of the triplet loss designs when combining it with the classification loss, including the similarity functions (\ieno, Euclidean, Cosine similarity), sampling mechanisms (batch hard mining or not). Note that for triplet loss, we use soft-margin triplet (see (\ref{eq:triplet-S})) which has been demonstrated to be better than triplet with hard margin \cite{hermans2017defense} (we have the similar observations).

Table \ref{tab:tri_cls} shows the results. We have the following observations. \textbf{1)} The joint use of triplet loss (Tri.) and classification loss (Cls.) achieves much better performance than using only one, even brings 3.2\% improvement in mAP on CUHK03(L). Without classification loss, triplet loss alone using cosine similarity suffers from difficulty in optimization and is easy to be trapped to local optimal. Thus, they are complementary in learning discriminative features. \textbf{2)} When Tri. and Cls. are jointly used, employing cosine similarity in general significantly outperforms employing (the negative of) Euclidean distance. Note that in the ReID inference, normalization on each sample is generally performed for the matching to exclude the interference of the amplitudes (energies) of sample features. Cosine similarity inherently evaluates the correlation of two features with energy normalized and it aligns better with ReID inference. \textbf{3)} When Tri. and Cls. are jointly used with cosine similarity, batch hard mining (which selects the hard positive and hard negative samples to form a triple) is inferior to the scheme without hard mining. That may be because the soft-margin enables the optimization of moderate hard triples and easy triples while hard mining could ignore those triplets. This phenomena is not observed for the Euclidean distance setting.

Hereafter, we refer to the scheme (last row in Table \ref{tab:tri_cls}) with the best combination of design choices as \emph{Baseline}.

\subsection{Effectiveness of Multi-class Classification Loss and Our MPN-tuple Loss}
\label{subsec:Effectiveness}

We validate the effectiveness of the vanilla N-tuple loss and our proposed Meta Prototypical N-tuple (MPN-tuple) loss for person ReID. The conventional classification loss is always used hereafter in considering its complementary role. Table \ref{tab:multi-class} shows the results. \emph{Baseline} refers to our obtained strong baseline (\ieno, the best one in Table. \ref{tab:tri_cls}). \emph{P2S Tri.} refers to point-to-set triplet loss wherein the prototypes (the average results over all instances of the same class) are taken as the reference nodes. This can reduce the number of triplets for calculating multi-class classification losses within a batch to $B=64$. We observe that \emph{P2S Tri. + Cls.} is slightly inferior to \emph{Baseline} but the number of formed triplets within a batch is smaller.

\begin{table*}[t]
  \centering
  \caption{Performance ($\%$) comparisons for triplet loss, multi-class classification (N-tuple loss), and our proposed MPN-tuple loss. \emph{Baseline} refers to our strong baseline (the best one in Table \ref{tab:tri_cls}). \emph{P2S Tri.+Cls.} refers to the use of point-to-set triplet loss and classification loss. \emph{\#Cls} denotes the number of classes.}
    \begin{tabular}{c|c|c|c|c|c|c|c|c|c}
    \Xhline{0.8pt}
    \multirow{2}[0]{*}{Loss} & \multirow{2}[0]{*}{\#Cls} & \multicolumn{2}{c|}{CUHK03(L)} & \multicolumn{2}{c|}{Market1501} & \multicolumn{2}{c|}{DukeMTMC} & \multicolumn{2}{c}{MSMT17} \\
    \cline{3-10}
          &       & R1   & \,\,mAP\,\,   & R1   & \,\,mAP\,\,   & R1   & \,\,mAP\,\,   & R1   & \,\,mAP\,\, \\
    \hline
    Baseline(Tri.+Cls.) & 2     & 81.8  & 78.2  & 94.7  & 87.3  & 88.9  & 78.3  & 79.8  & 56.2 \\
    P2S Tri.+Cls. & 2     & 81.1  & 77.8  & 94.8  & 86.7  & 88.5  & 78.1  & 80.0  & 55.6 \\
    \hline
    N-tuple+Cls. & 2     & 81.4  & 77.7  & 94.4  & 87.0  & 88.9  & 78.1  & 79.2  & 55.7 \\
    N-tuple+Cls. & 4     & 82.1  & 78.4  & 94.5  & 87.2  & 88.8  & 78.6  & 80.0  & 57.2 \\
    N-tuple+Cls. & 8     & 82.1  & 78.9  & 94.5  & 87.5  & 89.0  & 79.0  &  80.3 & 57.8 \\
    N-tuple+Cls. & 16     & 82.2  & 79.1  & 94.7  & 87.7  & 89.4  & 79.2  &  80.2 & 58.1 \\
    \hline
    PN-tuple+Cls. & 16    & 82.9  & 79.6  & 94.8  & 87.5  & 89.7  & 78.8  & 80.9  & 58.2 \\ 
    \multicolumn{1}{c|}{\textbf{MPN-tuple}+Cls.} & 16    & \textbf{84.4}  & \textbf{80.3}  & \textbf{95.3}  & \textbf{88.7}  & \textbf{89.5}  & \textbf{79.7}  & \textbf{82.2} & \textbf{60.1} \\
    \Xhline{0.8pt}
    \end{tabular}%
  \label{tab:multi-class}%
\end{table*}%

\begin{table*}[t]
  \centering
  \caption{Performance ($\%$) comparisons with the state of the art methods. Bold numbers denote the best performance and the numbers with underlines denote the second best ones.}
  \label{tab:SOTA}
  \newcommand{\tabincell}[2]{\begin{tabular}{@{}#1@{}}#2\end{tabular}}
    \begin{tabular}{c|c|c|c|c|c|c|c|c}
    \Xhline{0.8pt}
    \multirow{2}[0]{*}{Model} & \multicolumn{2}{c|}{CUHK03(L)} & \multicolumn{2}{c|}{Market1501} & \multicolumn{2}{c|}{DukeMTMC} & \multicolumn{2}{c}{MSMT17} \\
    \cline{2-9}
    \multicolumn{1}{c|}{} & R1 & mAP & R1 & mAP & R1 & mAP & R1 & mAP \\
    \hline
    \multicolumn{1}{l|}{HAP2S \cite{yu2018hard}} &    -   &    -   & 84.6  & 69.4  & 76.0  & 60.6  &    -   & - \\
    \multicolumn{1}{l|}{CE-FAT \cite{yuan2019defense}} &   -    & -    & 91.4  & 76.4  & 80.8  & 63.1  & 69.4  & 39.2 \\
    \multicolumn{1}{l|}{IDE \cite{zheng2017person}} & 43.8  & 38.9  & 85.3  & 68.5  & 73.2  & 52.8  &       &  \\
    \multicolumn{1}{l|}{IDO-Cls \cite{zhai2019defense}} & 62.8  & 56.7  & 93.9  & 80.5  &       &       &       &  \\
    \multicolumn{1}{l|}{Gp-reid \cite{almazan2018re}} &   -    &   -    & 92.2  & 81.2  & 85.2  & 72.8  &   -    &  - \\
    \multicolumn{1}{l|}{Bag of Tricks \cite{luo2019strong}} &  -     &    -   & 94.5  & 85.9  & 86.4  & 76.4  &  -   &  -\\
    \multicolumn{1}{l|}{IANet \cite{hou2019interaction}} &   -    &   -    & 94.4  & 83.1  & 87.1  & 73.4  & 75.5  & 46.8  \\
    \multicolumn{1}{l|}{HCTL \cite{zhao2020deep}} & - & - & 93.8 & 81.8 & 83.3 & 68.2 & 74.3 & 43.6 \\
    \multicolumn{1}{l|}{PCB+RPP \cite{sun2017beyond}} & 63.7  & 57.5  & 93.8  & 81.6  & 83.3  & 69.2  & 68.2  & 40.4  \\
    \multicolumn{1}{l|}{MGN \cite{wang2018learning}} & 68.0  & 67.4  & 95.7  & 86.9  & 88.7  & \underline{78.4}  &   -    &  \\
    \multicolumn{1}{l|}{DSA-reID \cite{zhang2019densely}} & 78.9  & 75.2  & 95.7  & 87.6  & 86.2  & 74.3  &    -   &  \\
    \multicolumn{1}{l|}{SAN \cite{jin2019semantics}} & 80.1  & 76.4  & \textbf{96.1}  & \underline{88.0}  & 87.9  & 75.5  &   79.2    & 55.7 \\
    \multicolumn{1}{l|}{MHN-6(PCB) (Chen et al. 2019)} & 77.2  & 72.4  & 95.1  & 85.0  & \underline{89.1}  & 77.2  &    -   & - \\
    \multicolumn{1}{l|}{BAT-net \cite{fang2019bilinear}} & 78.6  & 76.1  & 95.1  & 84.7  & 87.7  & 77.3  & 79.5  & 56.8  \\
    \multicolumn{1}{l|}{OSNet \cite{zhou2019omni}} &   -    &   -   &94.8  & 84.9  & 88.6  & 73.5  & 78.7  & 52.9  \\
    \multicolumn{1}{l|}{Mancs \cite{wang2018mancs}} & 69.0  & 63.9  & 93.1  & 82.3  & 84.9  & 71.8  &   -    &  -\\
    \multicolumn{1}{l|}{JDGL \cite{zheng2019joint}}  &   -    &  -  & 94.8  & 86.0  & 86.6  & 74.8  & 77.2  & 52.3  \\
    \multicolumn{1}{l|}{RGA-SC \cite{zhang2019relation}} & 80.4  & 76.5  & \underline{95.8}  & 88.1  & 86.1  & 74.9  & \underline{81.3}  & \underline{56.3}  \\
    \multicolumn{1}{l|}{GASM \cite{heguided}} & - & - & 95.3 & 84.7 & 88.3 & 74.4 & 79.5 & 52.5 \\
    \multicolumn{1}{l|}{FIDI \cite{yan2020beyond}} & 75.0 & 73.2 & 94.5 & 86.8 & 88.1 & 77.5 & - & - \\
    \hline
    \multicolumn{1}{l|}{\textbf{Baseline}(Cls.+Tri.)} & \underline{81.8}  & \underline{78.2}  & 94.7  & 87.3  & 88.7  & 78.3  & 79.8  & 56.2  \\
    \multicolumn{1}{l|}{\textbf{Ours}(Cls.+MPN-tuple)} & \textbf{84.4}  & \textbf{80.3}  & 95.3  & \textbf{88.7}  & \textbf{89.5}  & \textbf{79.7}  & \textbf{82.2} & \textbf{60.1} \\
    \bottomrule
    \end{tabular}
\end{table*}

\noindent\textbf{Influence of the Number of Classes in N-tuple Loss.}
We investigate the influence of the number of jointly considered classes (denoted by \# Cls) for a given query by increasing the $N$ in N-tuple loss. With the increase of $N$, the number of possible tuples $C_P^N \cdot K^N N(K-1)$ increases exponentially such that it quickly becomes intractable. We randomly sample $M$ tuples to calculate the losses. For fair comparison among schemes with different number of classes, we set $M=T$, where $T$ is the number of total triplets in a batch. We denote these schemes as \emph{N-tuple+Cls.}.

In Table \ref{tab:multi-class}, we observe that as the number of classes in \emph{N-tuple} loss increases, the person ReID performance in general increases. When the number of classes increases from 2 to 16, the mAP accuracy is improved by 1.4\%, 0.7\%, 1.1\%, and 2.4\% on CUHK03, Market1501, DukeMTMC, and MSMT17, respectively. Note that since the \emph{Baseline} is already very strong with superior performance, our gains on top of it can be considered as significant. The gradients of two different triplets may contradict each other as discussed in Section \ref{sec:introduction}. Although $P$ identities are involved in calculating the triplet loss at the batch level, it still does not allow valid interaction among different triplets. In N-tuple loss, as $N$ increases, more instances are jointly considered so that the corresponding instances are optimized towards the correct ranking with respect to the given query. Note that the performance of \emph{N-tuple + Cls.} with $N=2$ is lower than our \emph{Baseline~(Tri.+Cls.)} because the random sampling in N-tuple loss cannot assure a complete traversal over all triplets even though the number of samplings is the same as the number of all triplets.

\noindent\textbf{Effectiveness of Our Proposed MPN-tuple Loss.}
To reduce the number of possible tuples in N-tuple loss, we calculate the prototype of multiple instances of the same identity and take the calculated prototype as reference node instead of using sample instance. In this way, the number of possible tuples is reduced from $C_P^N \cdot K^N N(K-1)$ to $B\cdot C_P^N$. We denote such scheme as \emph{PN-tuple + Cls.}. Taking the extreme $N=16$ case as an example, the number of all possible tuples is reduced to 64 and we use the 64 tuples to calculate the losses. \emph{PN-tuple + Cls.} achieves similar performance as \emph{N-tuple + Cls.} when $N=16$ but avoids the experience of too many tuples in calculating the loss value.

\emph{MPN-tuple+Cls.} denotes our scheme where a meta learner (a mapping subnet) is introduced for inferring reference nodes. This enables a more general similarity metric for effective feature learning. We can see that \emph{MPN-tuple+Cls.} with $N=16$ achieves significant improvement over \emph{PN-tuple+Cls.}, \ieno, 0.7\%, 1.2\%, 0.9\%, and 1.9\% gain in mAP accuracy on four benchmark datasets, respectively. Thanks to the introduction of prototypes and the meta-learner, \emph{MPN-tuple+Cls.} achieves the best performance. Note that at the initial stage when the network has not been trained well, it is challenging for the meta learner to learn a good representation. Thus, we use three-stage training. In the first stage, we train the network with classification loss and PN-tuple loss for 360 epoches. In the second stage (361-480 epoches), we fix the  network and only train the meta-learner with MPN-tuple loss and the FC layer corresponding to the classification loss. In the third stage (480-600 epoches), we jointly finetune the entire network.

Note that it's desirable to involve as many classes as possible but impractical in the case when the number of classes is limited within a batch. We fix the batch size for fair comparison and convincing ablation study, because the learning for person ReID is sensitive to the batch size used. Thus, the number of classes is also limited. We set it to 16.

\subsection{Comparison with the State-of-the-Arts}

Table \ref{tab:SOTA} shows the comparison results with the state-of-the-art approaches. We group these approaches into two groups. The first group aims at designing strong baseline networks, including loss deigns and training tricks. In \cite{luo2019strong}, bag of tricks are collected and evaluated for person ReID and a strong baseline built based on ResNet-50 is provided. The other group of approaches focuses on special network designs for person ReID. Some approaches \cite{sun2017beyond,wang2018learning,chen2019mixed} ensemble the local region feature representations, while others introduce attention designs to focus on discriminative features \cite{wang2018mancs,fang2019bilinear,zhang2019relation}. Besides, there are some approaches exploiting auxiliary semantics (\egno, dense semantics \cite{zhang2019densely,jin2019semantics}) to address the misalignment (caused by the diverse viewpoints and poses).  

\begin{table*}[th]
  \centering
  \caption{Performance ($\%$) comparisons for triplet loss, multi-class classification (N-tuplet loss), and our proposed MPN-tuple loss on top of the \emph{Plain Baseline}. The symbol ``\cmark'' represents ``using IBN'' while ``\xmark'' represents ``all not using IBN".}
    \begin{tabular}{c|c|c|c|c|c|c|c|c|c|c|c}
    \Xhline{0.8pt}
    \multirow{2}[0]{*}{Model \& Loss} & \multirow{2}[0]{*}{\# Class} & \multirow{2}[0]{*}{\,\,IBN\,\,} & \multirow{2}[0]{*}{Resolution}  & \multicolumn{2}{c|}{CUHK03(L)} & \multicolumn{2}{c|}{Market1501} & \multicolumn{2}{c|}{DukeMTMC} & \multicolumn{2}{c}{MSMT17} \\
    \cline{5-12}
          &       &       &       & Rank-1   & \,\,mAP\,\,   & Rank-1   & \,\,mAP\,\,   & Rank-1   & \,\,mAP\,\,   & Rank-1   & \,\,mAP\,\, \\
    \hline
    \multicolumn{1}{l|}{(1) Baseline (Tri. + Cls.)} & 2     & \xmark  & $256 \times 128$ & 69.7  & 66.5  & 93.6  & 83.4  & 86.0  & 75.5  & 71.9  & 46.2 \\
    \hline 
    \multicolumn{1}{l|}{(2) N-tuplet + Cls.} & 2     & \xmark  & $256 \times 128$ & 69.3  & 66.1  & 94.2  & 84.2 & 85.9  & 75.5  & 71.5  & 46.3 \\
    \multicolumn{1}{l|}{(3) N-tuplet + Cls.} & 4     & \xmark  & $256 \times 128$ & 71.9  & 68.6  & 94.5  & 85.6 & 87.2  & 76.8  & 72.9  & 48.2 \\
    \multicolumn{1}{l|}{(4) N-tuplet + Cls.} & 8     & \xmark  & $256 \times 128$ & 74.9  & 70.3  & 94.7  & 86.1 & 87.3  & 76.9  & 73.7  & 49.3 \\
    \multicolumn{1}{l|}{(5) N-tuplet + Cls.} & 16    & \xmark  & $256 \times 128$ & 75.4  & 71.0  & 94.7  & 86.2 & 86.8  & 77.3  & 74.3  & 49.9 \\
    \hline
    \multicolumn{1}{l|}{(6) MPN-tuple + Cls.} & 16    & \xmark & $256 \times 128$ & 77.7  & 73.4  & 94.6  & 86.4   & 87.3  & 77.5  & 77.6  & 52.4 \\
    \Xhline{0.8pt}
    \end{tabular}
  \label{tab:ourloss}%
\end{table*}%

\begin{table*}[htbp]
  \centering
  \caption{An ablation study on the impact of Instance Normalization in backbone network, and the input image resolution. \emph{Baseline} refers to our baseline models which use triplet loss (denoted by ``Tri.'') and conventional classification loss (denoted by ``Cls.'') as supervision (see Table~\ref{tab:tri_cls}). ``\cmark'' represents ``using IBN'' while ``\xmark'' represents ``all not using IBN".}
    \begin{tabular}{c|c|c|c|c|c|c|c|c|c|c|c}
    \Xhline{0.8pt}
    \multirow{2}[0]{*}{Model \& Loss} & \multirow{2}[0]{*}{\# Class} & \multirow{2}[0]{*}{\,\,IBN\,\,} & \multirow{2}[0]{*}{Resolution}  & \multicolumn{2}{c|}{CUHK03(L)} & \multicolumn{2}{c|}{Market1501} & \multicolumn{2}{c|}{DukeMTMC} & \multicolumn{2}{c}{MSMT17} \\
    \cline{5-12}
          &       &       &       & Rank-1   & \,\,mAP\,\,   & Rank-1   & \,\,mAP\,\,   & Rank-1   & \,\,mAP\,\,   & Rank-1   & \,\,mAP\,\, \\
    \hline
    \multicolumn{1}{l|}{(1) Baseline (Tri. + Cls.)} & 2     & \xmark  & $256 \times 128$ & 69.7  & 66.5  & 93.6  & 83.4  & 86.0  & 75.5  & 71.9  & 46.2 \\
    \multicolumn{1}{l|}{(2) Baseline (Tri. + Cls.)} & 2    & \xmark  & $384 \times 128$ & 76.4  & 73.1  & 94.4  & 85.2  & 87.8  & 77.5  & 76.1  & 52.2 \\
    \multicolumn{1}{l|}{(3) Baseline (Tri. + Cls.)} & 2     & \cmark  & $384 \times 128$ & 81.8  & 78.2  & 94.7  & 87.3  & 88.9  & 78.3  & 79.8  & 56.2 \\
    \multicolumn{1}{l|}{(4) Ours (MPN-tuple + Cls.)} & 16  & \cmark  & $384 \times 128$ & \textbf{84.4}  & \textbf{80.3}  & \textbf{95.3}  & \textbf{88.7}  & \textbf{89.5}  & \textbf{79.7}  & \textbf{82.2} & \textbf{60.1} \\
    \Xhline{0.8pt}
    \end{tabular}
  \label{tab:ablation}%
\end{table*}%

Our study belongs to the first group. Thanks to the re-investigation on triplet loss design choices, we provide a strong baseline sdheme \emph{Baseline}, which jointly uses the soft-margin triplet loss (with cosine similarity, without hard mining) and classification loss (see Table \ref{tab:tri_cls} about the ablation study). We can see that our \emph{Baseline} achieves high performance, being superior or competitive to the state-of-the-art approaches. We denote our final scheme with the proposed MPN-tuple loss as \emph{MPN-tuple + Cls}. It achieves the best mAP accuracy on all these datasets and outperforms the \emph{Baseline} by a large margin, achieving 2.1\%, 1.4\%, 1.4\% and 3.9\% gain in mAP accuracy on the four datasets, respectively. Our design is simple yet effective. There is no increase in the computational complexity during the inference. We hope our scheme could serve as a strong baseline for the ReID community and inspire more novel designs on loss functions.

\subsection{Effectiveness of our Proposed Loss Designs on the Plain Baseline}
\label{subsec:LossDesign}

We have validated the effectiveness of N-tuple loss and our proposed Meta Prototypical N-tuplet (MPN-tuple) loss on top of our strong baseline scheme \emph{Baseline} in Section~\ref{subsec:Effectiveness}. Here, we further study their effectiveness on top of the \emph{Plain Baseline} (\ieno, the model (1) in Table~\ref{tab:ourloss}) which does not use IN and uses the low resolution setting. Table~\ref{tab:ourloss} shows the results. 

For model (2) to (5) in Table~\ref{tab:ourloss}, we can see that both the mAP and Rank-1 accuracy are consistently and significantly improved as the number of classes (denoted by ``\# Class") increases for the N-tuple loss. When the number of classes increases from 2 to 16, the mAP accuracy is improved by 4.9\%, 2.0\%, 1.8\% and 3.6\% on CUHK03, Market1501, DukeMTMC, and MSMT17 respectively, which is more significant than that on the strong baseline (see Table~\ref{tab:ablation}  uses IBN and a higher resolution). This demonstrates that the increase of the number of jointly considered classes in per-query optimization is very helpful for ReID, since such comparisons among more classes are more consistent with the ReID test/inference which is actually a retrieval/comparison process.

The model with the incorporation of our proposed MPN-tuple loss (\ieno, model (6)) significantly outperforms the \emph{Plain Baseline}, \ieno, model (1), by 6.9\%, 3.0\%, 2.0\% and 6.2\% on CUHK03, Market1501, DukeMTMC, and MSMT17, respectively. Our MPN-tuple loss enables stronger metric learning and is superior to N-tuplet loss.

\subsection{Influence of Input Resolution and IBN on Baseline}

We have conducted an empirical study of the effects of loss designs for person ReID baseline models (see Section~\ref{subsec:Effectiveness}). For those settings in Section~\ref{subsec:Effectiveness}, we use ResNet-50 backbone with Instance Normalization (IN) inserted (\ieno, IBN) \cite{pan2018two} and the input image resolution is 384$\times$128. Note that 384$\times$128 and 256$\times$128 are two commonly used input resolutions in person ReID literatures. Here, we study the influence of Instance Normalization (IN) and the input resolutions. Table~\ref{tab:ablation} shows the results. 

In Table~\ref{tab:ablation}, we refer to model (1) which does not use IBN and uses the low resolution setting as \emph{Plain Baseline} while referring model (3) that uses IBN and a higher resolution as \emph{Strong Baseline}. Comparing \emph{Plain Baseline} to \emph{Strong Baseline}, we find both factors bring significant improvements, wherein IBN enhances the generalization performance for the ``unseen'' testing identities, and higher resolutions can preserve more details of the input images. When compared with the \emph{Strong Baseline} (\ieno model (3) in Table~\ref{tab:ablation}), our proposed MPN-tuple loss brings 2.1\%, 1.4\%, 1.4\%, and 3.9\% improvements in mAP on CUHK03, Market1501, DukeMTMC, and MSMT17, respectively. This demonstrates the effectiveness of our MPN-tuple loss. Note that the gain of MPN-tuple loss on \emph{Strong Baseline} (see Table ~\ref{tab:ablation}) is smaller than that on \emph{Plain Baseline} (see Table~\ref{tab:ourloss}), which is because the stronger the baseline, the harder it is to obtain additional gain based on this baseline.

\begin{figure*}[t]
	\begin{center}
 		\includegraphics[width=1.0\linewidth]{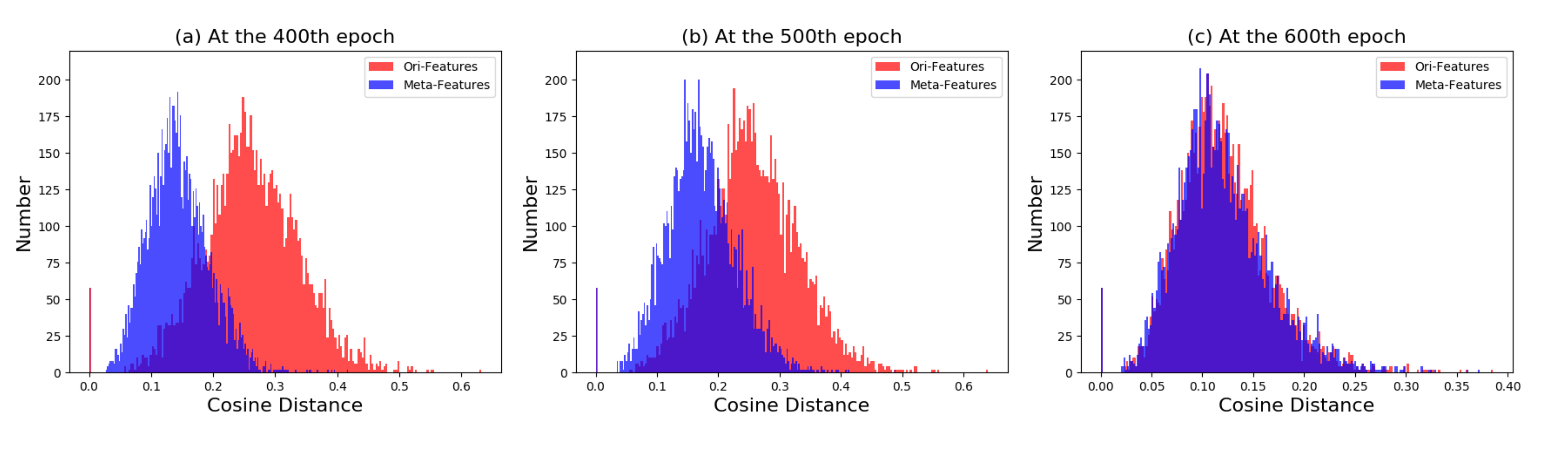}
	\end{center}
	\caption{Histograms of distances of positive sample pairs (\ieno, the pairs of the same identity). Red: within-class distances between \emph{original features} of two samples (\ieno, the features before the mapping via the meta learner (a subnet)). Blue: within-class distances between \emph{meta-features} of two samples (\ieno, the output features of the meta learner).}
	\label{fig:histogram}
\end{figure*}

\subsection{Analysis of Features Learned from Our MPN-tuple Loss}
\label{sec:visualization}

In our proposed MPN-tuple loss, we adopt a trainable meta-learner to predict the category-specific reference node as a refined representation from each instance in a mini-batch. The meta-learner is end-to-end trained. Minimizing the loss on the ``refined" feature vectors (after meta-learner) will further derive the optimization of the input features of the meta-learner through end-to-end training. This is prone to enforce closer distances between the anchor and positive samples in the feature space and farther distances between the anchor and negative samples to ease the optimization. To demonstrate this, we visualize the the histogram of distances of positive sample pairs as the training process goes in Fig.~\ref{fig:histogram}.

Based on our model which is trained with \emph{MPN-tuple+Cls.} losses on the training set of Market1501, we visualize the distribution of the feature distances of positive (\ieno, the same identity) sample pairs at the 400$^{th}$ epoch, the 500$^{th}$ epoch, and the 600$^{th}$ epoch in Fig.~\ref{fig:histogram} (a), (b), and (c), respectively. We refer to the features before the subnet mapping (meta learner) as \emph{original features} $\boldsymbol{x}$ (namely \emph{Ori-Features}), and the features after the subnet mapping as \emph{meta-features} $\phi(\boldsymbol{x})$ (namely \emph{Meta-Features}). For the red/blue histogram marked by \emph{Ori-Features}/\emph{Meta-Features}, we obtain the feature distance of a sample pair (sample A, sample B) by calculating the Cosine distance (\ieno, One minus the Cosine similarity) between the sample A and sample B by using \emph{original features}/\emph{meta-features}.

We can observe that at the early training epochs (\egno, the 400$^{th}$ epoch shown in Fig.~\ref{fig:histogram}(a)) after the proposed MPN-tuple loss is employed, the distances of positive sample pairs calculated using meta-features are smaller than those calculated using original features, indicating the within-class similarity based on the meta-features is higher than that based on original features. The learned meta-features thus can be viewed as predictions for the identity representations from the original features by exploiting the patterns/correlations among different feature dimensions. Within the proposed MPN-tuple loss, we calculate the Cosine similarity between the original feature of one sample $\boldsymbol{x_a}$ and the meta feature $\phi(\boldsymbol{x_b})$ of another sample $\boldsymbol{x_b}$, which plays a equivalent role of learning a more general similarity metric between $\boldsymbol{x_a}$ and $\boldsymbol{x_b}$ \cite{qiao2018few}. As the training goes on, optimized by the MPN-tuple loss, the network becomes better under this learned similarity metric and makes original features approach the corresponding predicted identity representations such that the within-class similarity are further enhanced during training. We observe that the distances of positive pairs using \emph{original features} become smaller and smaller. In the final, the discrepancy of these two distributions (red histogram and blue histogram) are quite small after 600 epochs (see Fig.~\ref{fig:histogram}(b) and (c)). Thus, during the inference, we can get rid of that mapping subnet and directly use original features as the ReID feature $\boldsymbol{x}$ for matching (the difference is marginal: $<$0.4\%).

\begin{table*}[h]
  \centering
  \caption{Performance (\%) comparisons on visible-infrared person ReID. We adopt different test modes on RegDB, \ieno, ``Visible to Thermal'' and ``Thermal to Visible'', where ``Visible to Thermal'' refers to that visible images are taken as the query set while thermal images are taken as the gallery set, and so on. There are also two different modes for the single-shot evaluation on SYSU-MM01, \ieno, ``All Search'' mode and ``Single Search'' mode. For ``All Search'' mode, all images are used. And for the ``Indoor Search'' mode, only indoor images from 1st, 2nd, 3rd, 6th cameras are used. }
    \begin{tabular}{c|c|c|c|c|c|c|c|c}
    \Xhline{0.8pt}
    \multirow{3}[0]{*}{Loss} & \multicolumn{4}{c|}{RegDB}     & \multicolumn{4}{c}{SYSU-MM01} \\
    \cline{2-9}
    & \multicolumn{2}{c|}{Visible to Thermal} & \multicolumn{2}{c|}{Thermal to Visible} & \multicolumn{2}{c|}{All Search} & \multicolumn{2}{c}{Indoor Search} \\
    \cline{2-9}
    & Rank-1 & mAP & Rank-1 & mAP & Rank-1 & mAP & Rank-1 & mAP \\
    \hline
    \multicolumn{1}{l|}{Cls. + Tri.} & 52.7  & 48.0 & 58.9 & 55.5 & 36.2  & 37.7  & 49.7 & 56.3 \\
    \multicolumn{1}{l|}{Cls. + TriHard.} & 68.3  & 62.9 & 67.7  & 62.1 & 43.2  & 42.4  & 51.1 & 58.1 \\
    \multicolumn{1}{l|}{Cls. + MPN-tuple} & 70.6  & 69.0 & 69.8 & 68.7 & 47.9  & 47.1  & 57.2 & 65.1 \\
    \Xhline{0.8pt}
    \end{tabular}%
  \label{tab:visible-infrared}%
\end{table*}%

\begin{table*}[t]
  \centering
  \caption{Performance (\%) comparisons on cloth-changing person ReID. On LTCC dataset, ``Standard'' refers to the evaluation setting where the images of the same identity and camera view are discarded during testing. And ``Cloth-changing'' denotes that the images with the same identity, camera view and clothes are discarded during testing. On PRCC dataset, ``Cross-clothes'' refers to the person matching with cloth changes whereas ''Same-clothes'' testing has no cloth changes.}
    \begin{tabular}{c|c|c|c|c|c|c|c|c|c|c}
    \Xhline{0.8pt}
    \multirow{3}[0]{*}{Loss} & \multicolumn{4}{c|}{LTCC}     & \multicolumn{6}{c}{PRCC} \\
    \cline{2-11}
    & \multicolumn{2}{c|}{Standard} & \multicolumn{2}{c|}{Cloth-changing} & \multicolumn{3}{c|}{Cross-Clothes} & \multicolumn{3}{c}{Same-Clothes} \\
    \cline{2-11}
    & Rank-1 & mAP & Rank-1 & mAP & Rank-1 & Rank-10 & Rank-20 & Rank-1 & Rank-10 & Rank-20 \\
    \hline
    \multicolumn{1}{l|}{Cls. + Tri.} & 58.5 & 27.3 & 39.6 & 11.3 & 26.2 & 35.9 & 40.0 & 80.5 & 87.4 & 90.0 \\
    \multicolumn{1}{l|}{Cls. + TriHard.} & 55.4 & 23.7 & 49.5 & 14.0 & 20.8 & 29.6 & 33.0 & 84.7 & 89.8 & 91.8 \\
    \multicolumn{1}{l|}{Cls. + MPN-tuple} & 59.8 & 29.8 & 53.0 & 21.6 & 35.2 & 42.1 & 45.0 & 88.2 & 93.1 & 94.8 \\
    \Xhline{0.8pt}
    \end{tabular}
  \label{tab:cloth-changing}%
\end{table*}%

\subsection{Effectiveness and Robustness Evaluation on Visible-Infrared Person ReID}

\noindent\textbf{Experimental Configurations.} For the visible-infrared person ReID, we evaluate the effectiveness of our proposed method by performing experimental comparisons on the commonly used datasets SYSU-MM01 \cite{wu2017rgb} and RegDB \cite{nguyen2017person}. In each mini-batch, we randomly sample 64 images with 16 identities where there are 4 RGB images and 4 thermal images for each identities. Empirically, we adopt SGD optimizer to train the entire model for 70 epochs. The initial learning rate is 0.1 and the learning rate decay is performed after the $40^{th}$ and the $50^{th}$ epoch. Other configurations are kept the same as the description in the Section~\ref{subsec:implementations}.

We adopt two different evaluation modes on RegDB \cite{nguyen2017person} and SYSU-MM01 \cite{wu2017rgb}, respectively. For RegDB, ``Visible to Thermal'' refers to that visible images are taken as the query set while thermal images are taken as the gallery set, and so on. For SYSU-MM01, ``All Search'' mode means that all images are used for testing. And under the ``Indoor Search'' mode, only indoor images from 1st, 2nd, 3rd, 6th cameras are used. The table~\ref{tab:visible-infrared} shows the ablation study results on visible-infrared person ReID. Our proposed MPN-tuple loss outperforms the triplet loss with hard mining (namely ``TriHard.'') by 2.3\%/6.1\% and 2.1\%/6.6\% for the ``Visible to Thermal'' and ``Thermal to Visible'' evaluation settings on the RegDB dataset and 4.7\%/4.7\% for 6.1\%/7.0\% on the SYSU-MM01 dataset in Rank-1/mAP, respectively. This experimental result shows that our proposed MPN-tuple loss has significant improvements compared to the triplet loss, indicating that MPN-tuple loss is also effective and robust for visible-infrared person ReID task.

\subsection{Effectiveness and Robustness Evaluation on Cloth-changing Person ReID}

For the cloth-changing person ReID, we evaluate the effectiveness of our proposed method by performing experimental comparisons on two most commonly used datasets, \ieno, PRCC \cite{yang2019person} and LTCC \cite{qian2020long}. In each mini-batch, we randomly sample 16 identities where each identity include 4 randomly sampled images, resulting in a batch size of 64. We adopt Adam optimizer to train the entire network for 70 epochs. During the training, we first warm up with a linear growth learning rate from $4 \times 10^{-6}$ to $4 \times 10^{-4}$ for 10 epochs and the learning rate is decayed by a factor of 0.5 for every 20 epochs. Other configurations are kept the same as the description in the Section~\ref{subsec:implementations}.

The table~\ref{tab:cloth-changing} shows performance comparisons for effectiveness evaluation. On the dataset LTCC, our proposed MPN-tuple loss achieves the Rank-1/mAP improvements of 4.4\%/6.1\% and 3.5\%/7.6\% in the ``Standard'' and ``Cloth-changing'' evaluation scenarios respectively compared to the triplet loss with hard mining (namely ``TriHard.''). On the dateset PRCC, MPN-tuple loss outperforms ``TriHard.'' by 14.4\% and 3.5\% in the Rank-1 accuracy under the ``Cross-clothes'' and ``Same-Clothes'' evaluation scenarios respectively. This experimental result verify the robustness of our proposed MPN-tuple loss on cloth-changing person ReID.

\section{Conclusions}
In this paper, we reformulate prevalent loss designs (triplet loss and classification loss) under a unified form and analyze their inherent limitations for person ReID. The triplet loss can be viewed as a two-class classification. There is a lack of valid interaction between different triplets and their optimization directions may contradict each other. The classification loss optimizes the similarity/distance between instances and parameter-based category centers, which enables stable global scope optimization but does not align well with the retrieval-based person ReID inference. Furthermore, we point out that N-tuple loss can provide more consistent optimization between training and testing but is under-explored for ReID task. Moreover, we introduce MPN-tuple loss which uses a meta learner to learn better references nodes (\ieno, better classifier) for more effective metric learning. The scheme powered by our proposed MPN-tuple loss achieves the state-of-the-art performance. Besides, we further verify the effectiveness and robustness of our proposed MPN-tuple loss on visible-infrared and cloth-changing person ReID. And we believe it has the potentials to be applied to other vision tasks.
We hope that in the future the ReID community will build on top of our strong baseline to investigate more novel loss designs.





\ifCLASSOPTIONcaptionsoff
  \newpage
\fi



%


\bibliographystyle{IEEEtran}
\bibliography{main}

%

\begin{IEEEbiography}[{\includegraphics[width=1in,height=1.25in,clip,keepaspectratio]{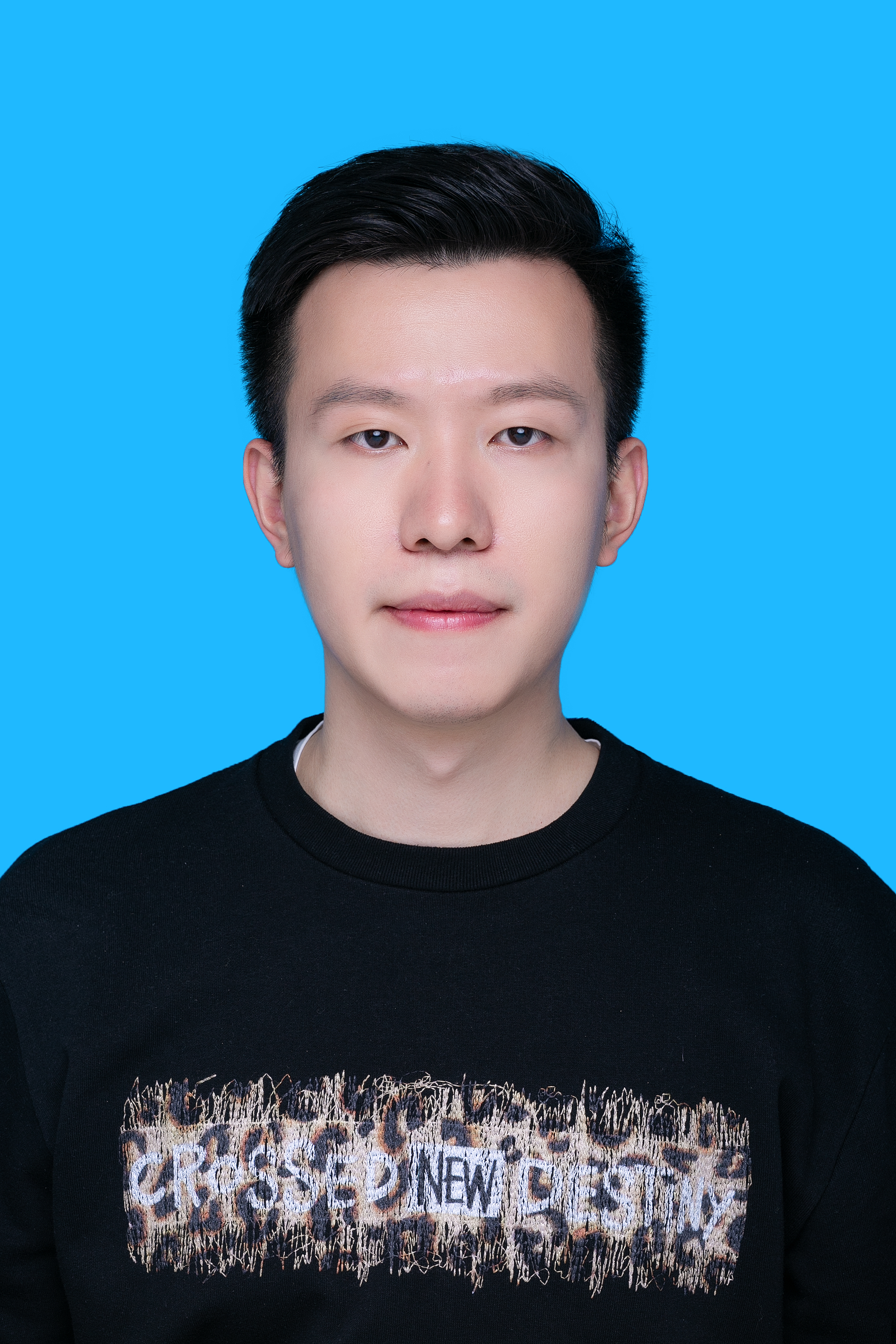}}]{Zhizheng Zhang}
received the B.S. degree from University of Electronic Science and Technology of China in 2016. He received the Ph.D. degree from University of Science and Technology of China in 2021. He joined Microsoft Research in June 2021 and is now a researcher at Microsoft Research Asia. His research interests include person re-identification, image/video compression, few-shot learning and domain generalization/adaptation.
\end{IEEEbiography}

\begin{IEEEbiography}[{\includegraphics[width=1in,height=1.25in,clip,keepaspectratio]{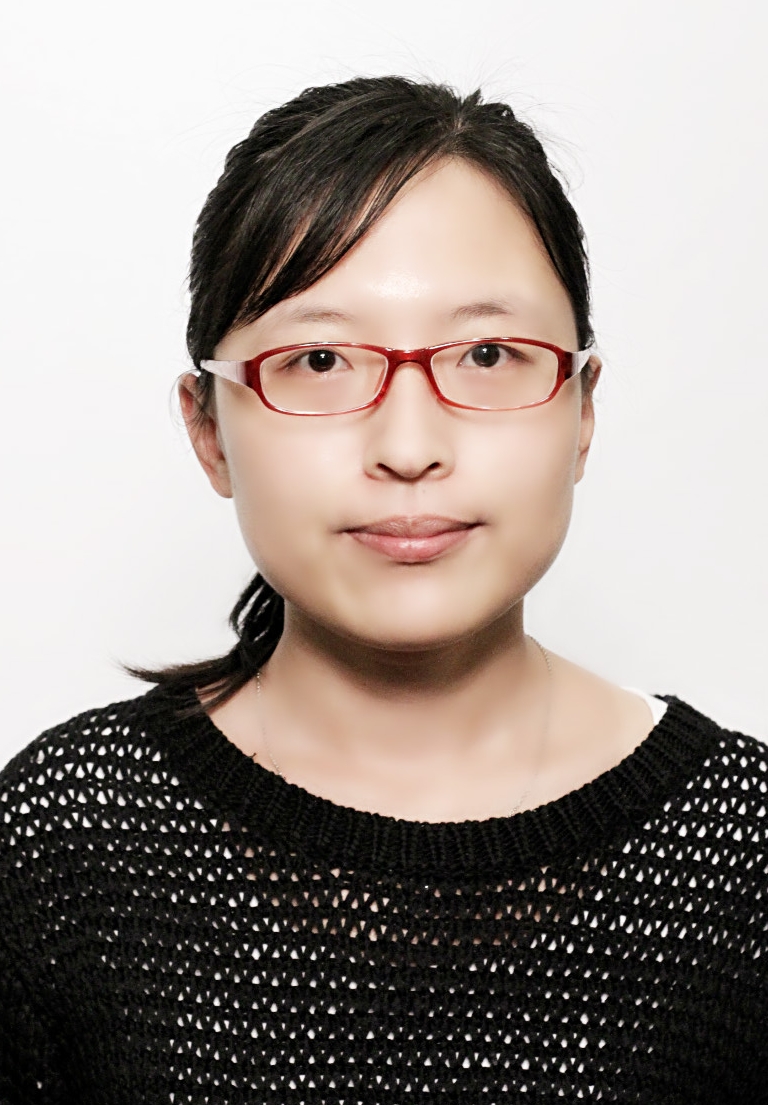}}]{Cuiling Lan}
received the B.S. degree in electrical engineering and the Ph.D. degree in intelligent information processing from Xidian University, Xi’an, China, in 2008 and 2014, respectively. She joined Microsoft Research Asia, Beijing, China, in 2014. Her current research interests include computer vision problems related to pose estimation, action recognition, person/ vehicle re-identification, domain generalization/adaptation.
\end{IEEEbiography}

\begin{IEEEbiography}[{\includegraphics[width=1in,height=1.25in,clip,keepaspectratio]{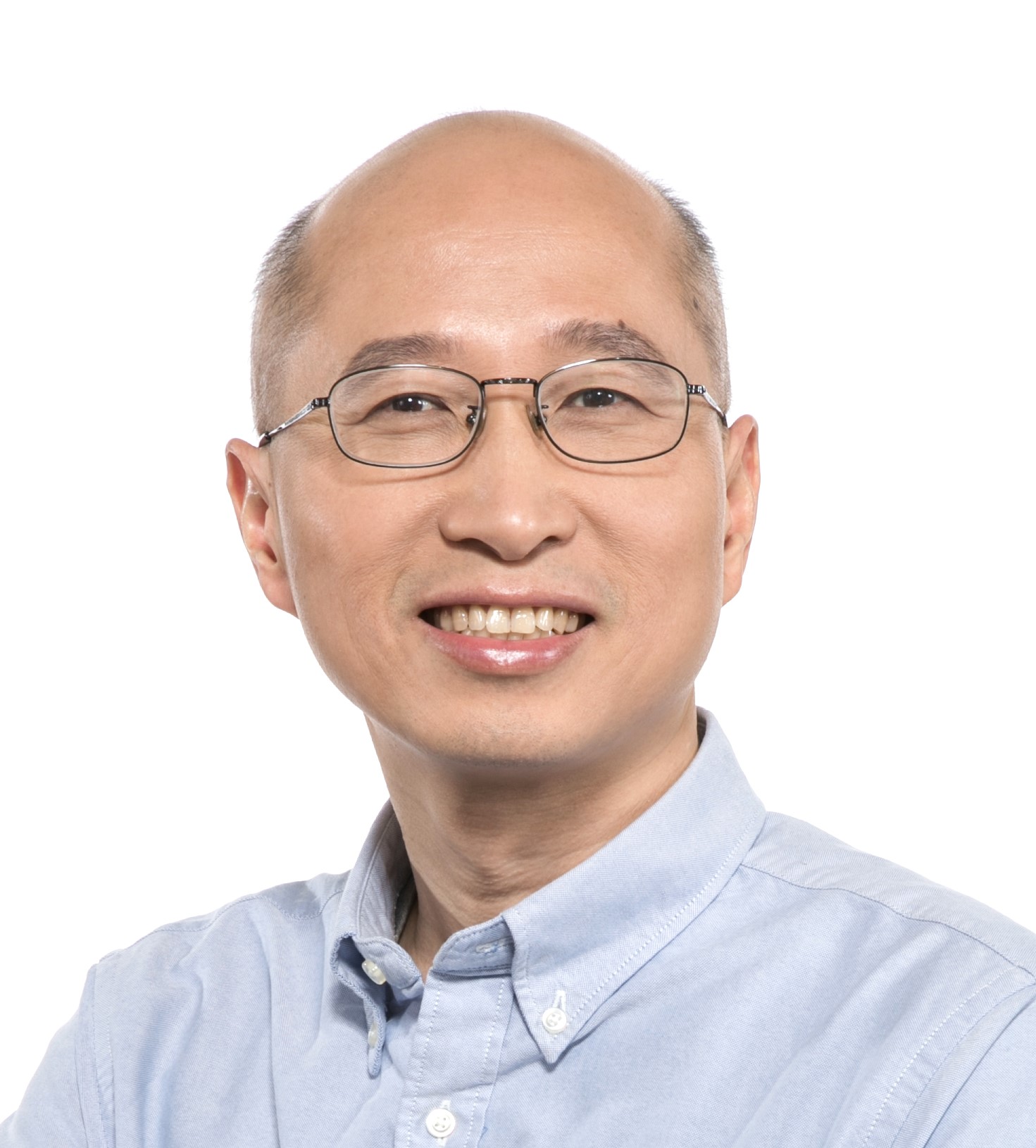}}]{Wenjun Zeng} 
(M'97-SM'03-F'12) is a Sr. Principal Research Manager and a member of the senior leadership team at Microsoft Research Asia. He has been leading the video analytics research empowering the Microsoft Cognitive Services, Azure Media Analytics Services, Office, Dynamics, and Windows Machine Learning since 2014. He was with Univ. of Missouri from 2003 to 2016, most recently as a Full Professor. Prior to that, he had worked for PacketVideo Corp., Sharp Labs of America, Bell Labs, and Panasonic Technology. Wenjun has contributed significantly to the development of international standards (ISO MPEG, JPEG2000, and OMA). He received his B.E., M.S., and Ph.D. degrees from Tsinghua Univ., the Univ. of Notre Dame, and Princeton Univ., respectively. His current research interests include mobile-cloud media computing, computer vision, and multimedia communications and security. He is on the Editorial Board of International Journal of Computer Vision. He was an Associate Editor-in-Chief of IEEE Multimedia Magazine, and was an AE of IEEE Trans. on Circuits $\&$ Systems for Video Technology (TCSVT), IEEE Trans. on Info. Forensics $\&$ Security, and IEEE Trans. on Multimedia (TMM). He was on the Steering Committee of IEEE Trans. on Mobile Computing and IEEE TMM. He served as the Steering Committee Chair of IEEE ICME in 2010 and 2011, and has served as the General Chair or TPC Chair for several IEEE conferences (e.g., ICME'2018, ICIP'2017). He was the recipient of several best paper awards. He is a Fellow of the IEEE.
\end{IEEEbiography}

\begin{IEEEbiography}[{\includegraphics[width=1in,height=1.25in,clip,keepaspectratio]{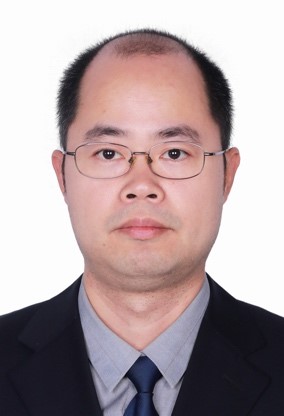}}]{Zhibo Chen} (M’01-SM’11) received the B. Sc., and Ph.D. degree from Department of Electrical Engineering Tsinghua University in 1998 and 2003, respectively. He is now a professor at University of Science and Technology of China. His research interests include image and video compression, visual quality of experience assessment, immersive media computing and intelligent media computing. He has more than 150 publications and more than 50 granted EU and US patent applications. He is IEEE senior member, Secretary/Chair-Elect of IEEE Visual Signal Processing and Communications Committee, and member of IEEE Multimedia System and Applications Committee. He was TPC chair of IEEE PCS 2019 and organization committee member of ICIP 2017 and ICME 2013, served as TPC member in IEEE ISCAS and IEEE VCIP.
\end{IEEEbiography}

\begin{IEEEbiography}[{\includegraphics[width=1in,height=1.25in,clip,keepaspectratio]{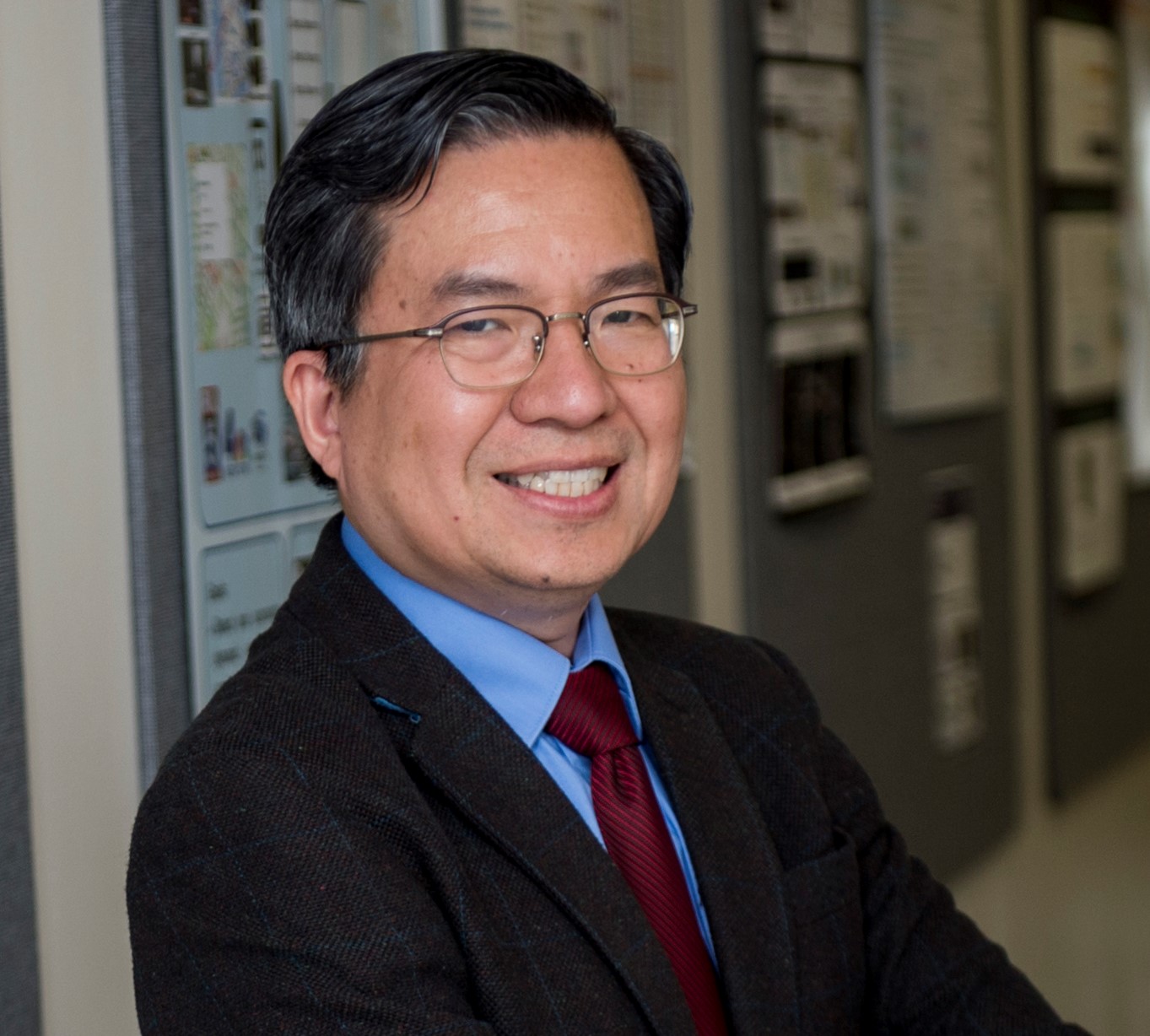}}]{Shih-Fu Chang}'s research is focused on computer vision, machine learning, and multimodal knowledge extraction. He received the IEEE Signal Processing Society Technical Achievement Award, ACM SIGMM Technical Achievement Award, the Honorary Doctorate from the University of Amsterdam, and the IEEE Kiyo Tomiyasu Award. He has been Interim Dean of Columbia Engineering (since 2021), an Amazon Scholar, a Fellow of AAAS, ACM, and IEEE, and an elected member of Academia Sinica.
\end{IEEEbiography}







\end{document}